\documentclass[conference]{IEEEtran}
\IEEEoverridecommandlockouts
\usepackage{cite}
\usepackage{url}
\usepackage{amsmath,amssymb,amsfonts}
\usepackage{algorithmic}
\usepackage{graphicx}
\usepackage{textcomp}
\usepackage{xcolor}
\usepackage{subcaption}
\usepackage{multirow}
\usepackage{enumitem} 
\usepackage{tikz}
\usepackage{makecell}
\def\BibTeX{{\rm B\kern-.05em{\sc i\kern-.025em b}\kern-.08em
    T\kern-.1667em\lower.7ex\hbox{E}\kern-.125emX}}
    
\begin{document}

\title{Deceptive Beauty: Evaluating the Impact of Beauty Filters on Deepfake and Morphing Attack Detection
}

\author{\IEEEauthorblockN{Sara Concas, Simone Maurizio La Cava, Andrea Panzino, Giulia Orr\'u, Ester Masala, Gian Luca Marcialis}
\IEEEauthorblockA{University of Cagliari, Piazza d'Armi I - 09123 Cagliari (Italy), e-mail: \\
\{sara.concas90c, simonem.lac, andrea.panzino, giulia.orru, marcialis\}@unica.it}}

\newcommand\copyrighttext{%
\footnotesize \textcopyright 2025 IEEE. Personal use of this material is permitted.
Permission from IEEE must be obtained for all other uses, in any current or future
media, including reprinting/republishing this material for advertising or promotional
purposes, creating new collective works, for resale or redistribution to servers or
lists, or reuse of any copyrighted component of this work in other works.}
\newcommand\copyrightnotice{%
\begin{tikzpicture}[remember picture,overlay]
\node[anchor=south,yshift=10pt] at (current page.south) {\fbox{\parbox{\dimexpr\textwidth-\fboxsep-\fboxrule\relax}{\copyrighttext}}};
\end{tikzpicture}%
}

\maketitle
\copyrightnotice

\begin{abstract}
Digital beautification through social media filters has become increasingly popular, raising concerns about the reliability of facial images and videos and the effectiveness of automated face analysis. This issue is particularly critical for presentation attack detectors, systems aiming at distinguishing between genuine and manipulated data, especially in cases involving deepfakes and morphing attacks designed to deceive humans and automated facial recognition.
This study examines whether beauty filters impact the performance of deepfake and morphing attack detectors. We conduct a comprehensive analysis, evaluating multiple state-of-the-art detectors on benchmark datasets before and after applying various beauty filters. Our findings reveal performance degradation, highlighting vulnerabilities introduced by facial enhancements and underscoring the need for robust detection models resilient to such alterations. 
\end{abstract}


\section{Introduction}
\begin{figure}[t]
    \centering
    \subfloat[Original]{%
        \includegraphics[width=0.27\columnwidth]{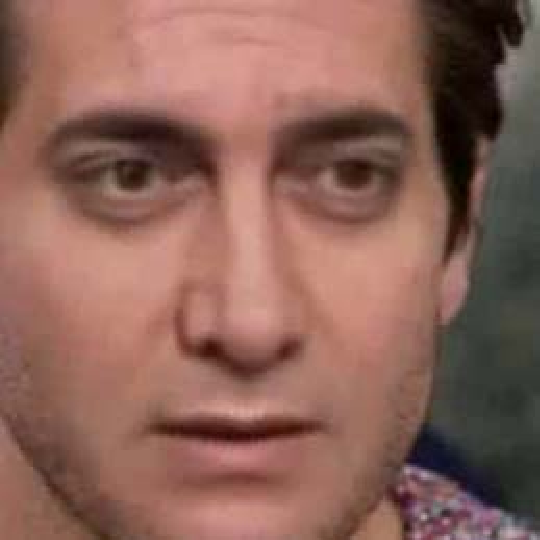}}
    \hfill
    \subfloat[$c=3\%$]{%
        \includegraphics[width=0.27\columnwidth]{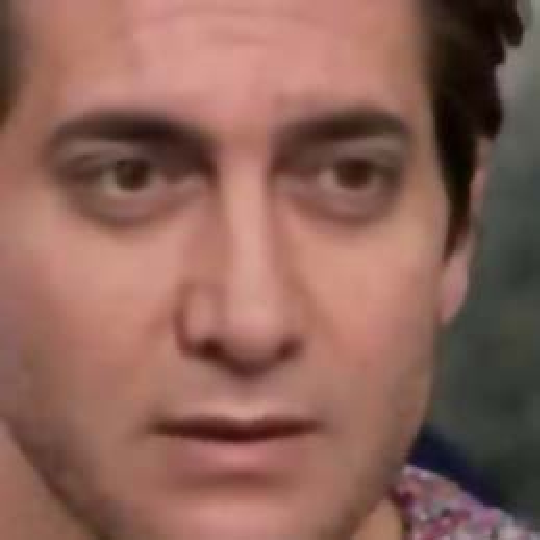}}
    \hfill
    \subfloat[$c=5\%$]{%
        \includegraphics[width=0.27\columnwidth]{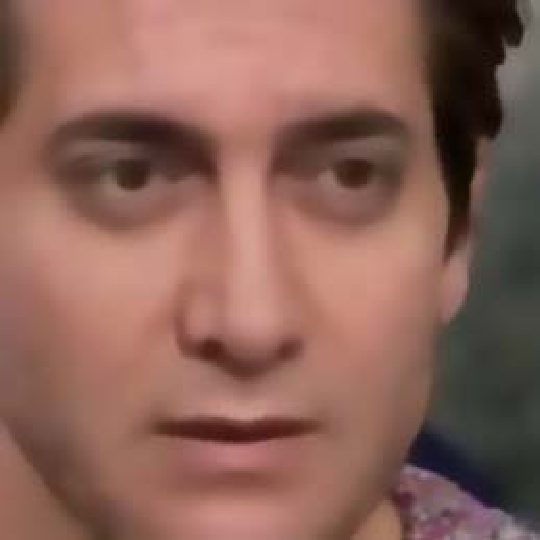}}

    \par\smallskip

    \subfloat[Original]{%
        \includegraphics[width=0.27\columnwidth]{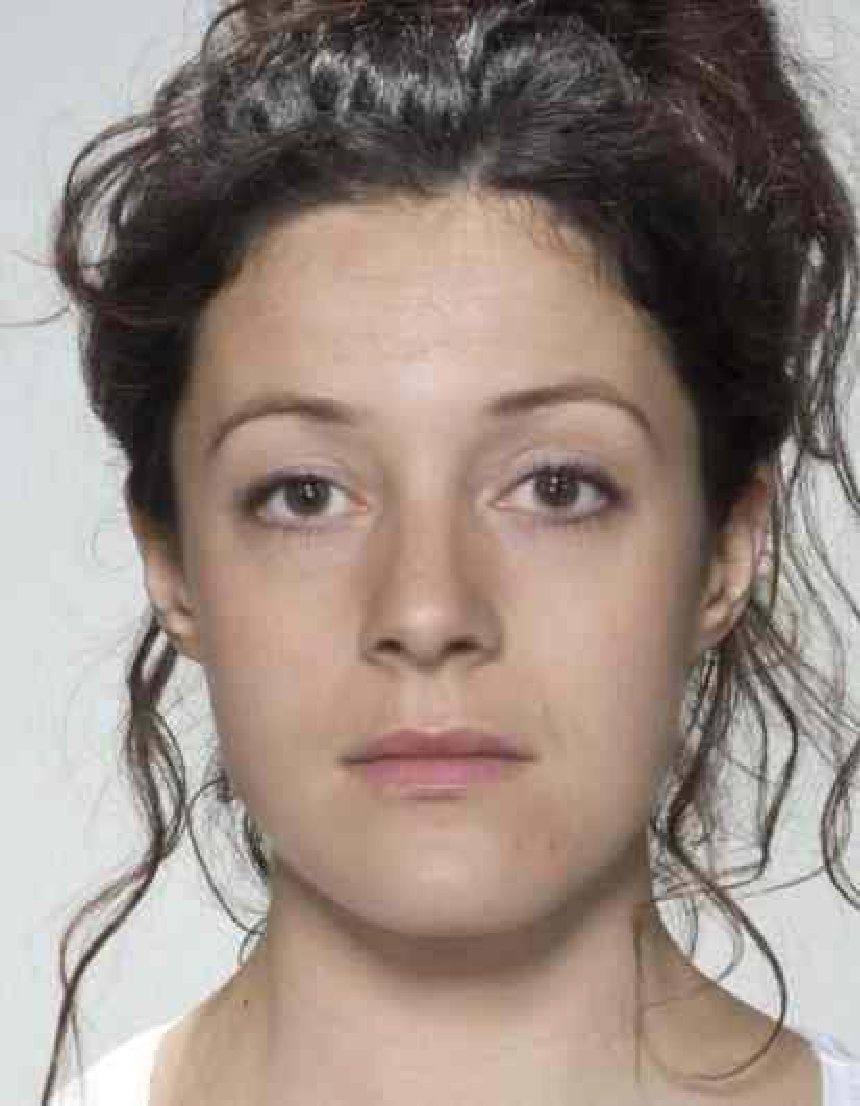}}
    \hfill
    \subfloat[$c=3\%$]{%
        \includegraphics[width=0.27\columnwidth]{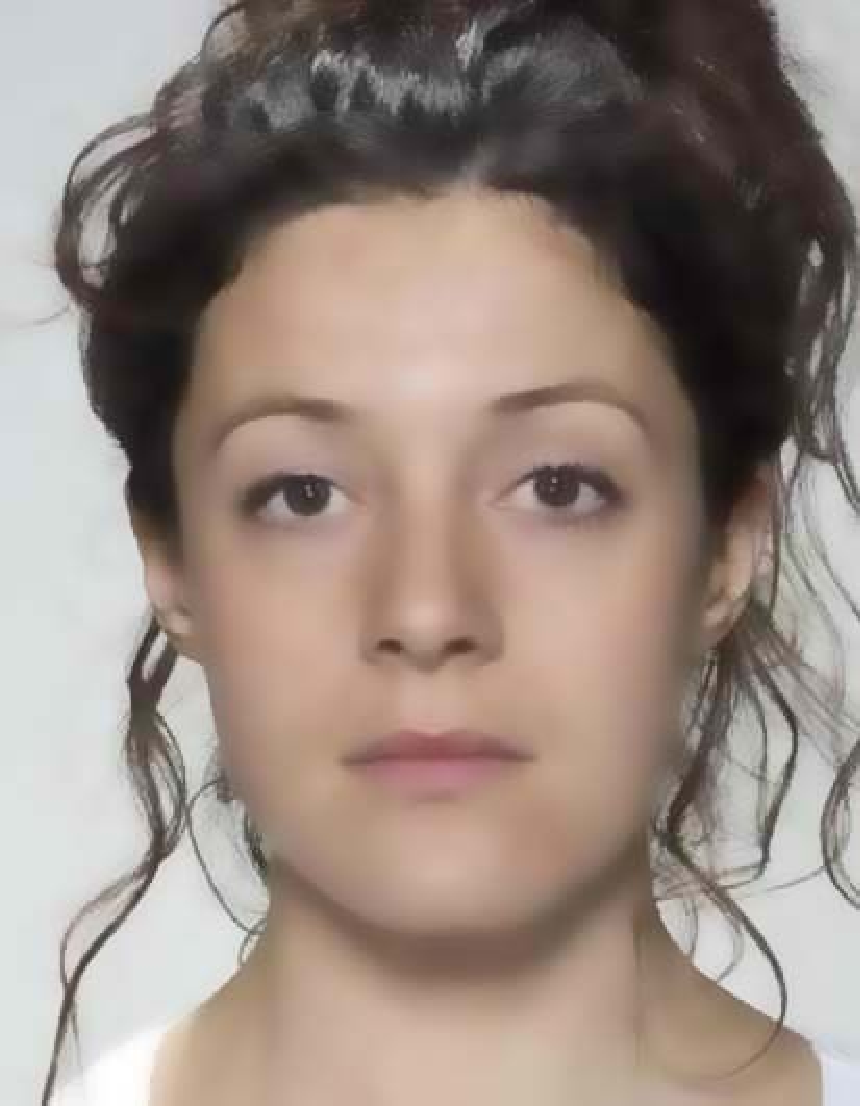}}
    \hfill
    \subfloat[$c=5\%$]{%
        \includegraphics[width=0.27\columnwidth]{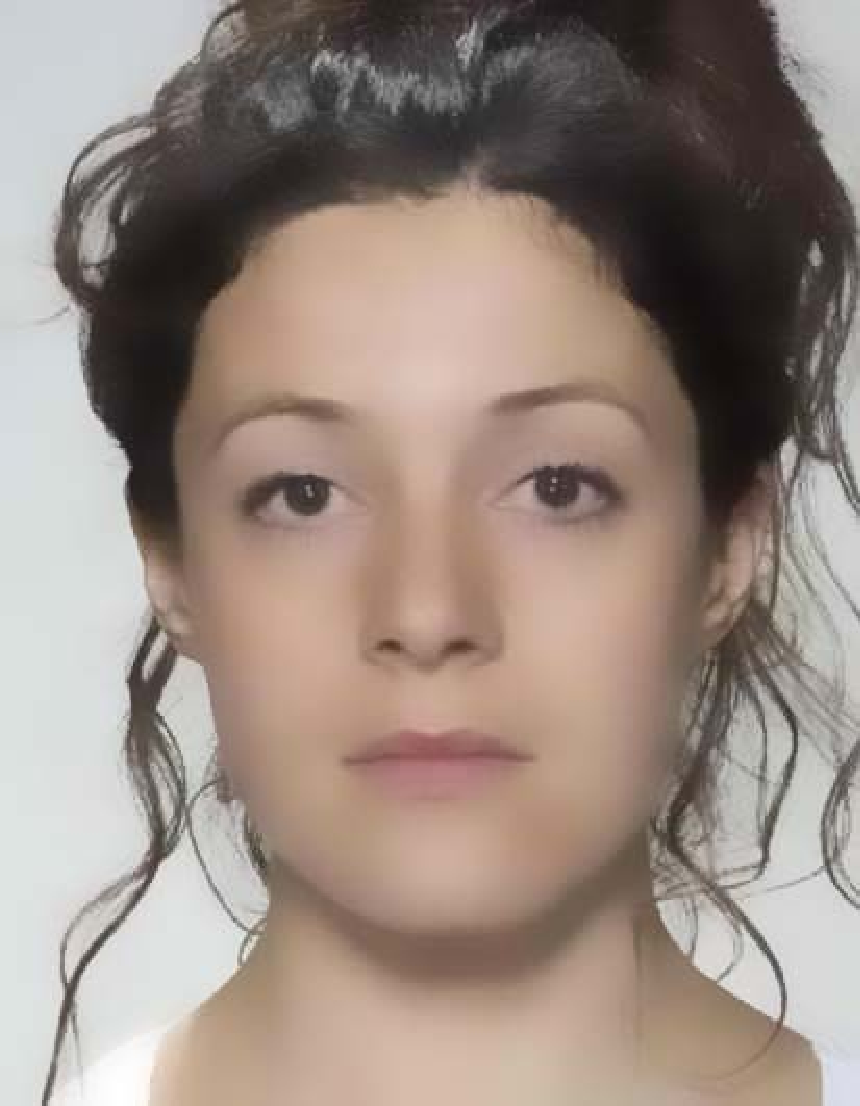}}

    \caption{Effect of two different values of the application radius of the smoothing filter using a smoothing radius $c$ (the percentage of the face height). The first row shows a deepfake example from~\cite{Celeb_DF_cvpr20}, the second a morph from~\cite{neubert2018extended}.}
    \label{fig:filters}
\end{figure}

The rise of social media and mobile technology has transformed digital content creation, making video and image sharing a daily activity for millions of users. Among the most prevalent trends is applying beauty filters, which digitally enhance facial features to align with evolving beauty standards \cite{javornik2022lies, yang2020selfie}. These tools are designed to automatically alter various facial features, such as skin, eyes, and lips, even with minimal user expertise. For instance, beauty filters are widely used for skin smoothing (Figure \ref{fig:filters}).

Although these filters offer aesthetic appeal and entertainment value, they introduce subtle yet impactful alterations that raise significant concerns regarding the authenticity and integrity of facial images and videos \cite{mirabet2022impact}. For instance, these filters can degrade the reliability of automated facial analysis technologies, including biometric authentication and identity verification systems, which rely on accurate and unaltered visual data \cite{mirabet2022impact, hedman2022effect}. 

This issue is particularly relevant in security applications, where distinguishing between genuine and manipulated facial data is critical. With the increasing prevalence of deepfake and morphing attacks, techniques capable of altering or blending identities to deceive recognition algorithms \cite{yu2021survey, ferrara2014magic}, the presence of beauty filters may introduce further challenges to the detection of manipulated data \cite{libourel2024case}. Beauty filters could be used for a dual purpose: i) bona fide users could naively use them simply to improve their appearance; ii) impostors could use them to hide manipulation artifacts. For this reason, assessing the robustness of detection systems against beautification practices could be critical to ensure the reliability required in security and digital forensic application scenarios in which these are typically employed \cite{hill2022police, jain2021biometrics, la20233d,venkatesh2021face}. While the research community has begun to explore the impact of beautification filters on biometric systems \cite{libourel2024case}, their effects on integrity detection remains largely unexplored, particularly concerning the impact of different facial manipulation technologies (e.i, face swap, morph attacks) on individual systems.

To address this gap, our study investigates whether the application of beauty filters affects the performance of deepfake and morphing attack detectors. We extensively evaluate state-of-the-art detection models on benchmark datasets, analyzing their robustness before and after applying beauty filters. In particular, our contributions are the following: (i) experiments on benchmark datasets, applying beauty filters to analyze their effect on the performance of two different state-of-the-art detectors, (ii) analyzing how increasing levels of facial enhancement affect the systems' performance, (iii) examination of the performance in various scenarios, jointly and separately considering the potential application of beautification filters on the real images (i.e., to improve the appearance) and on the fake images (i.e., as an attempt to deceive manipulation attack detectors).

The rest of the manuscript is organized as follows. Section \ref{sec:related} provides an overview of deepfakes and morphing attacks detection. Section \ref{sec:experimental} provides the experimental protocol considered to analyze the impact of beautification filters on the detection of these facial manipulations. Finally, results are reported in Section \ref{sec:results}, and conclusions are drawn in Section \ref{sec:conclusions}.

\section{Related Work}\label{sec:related}
Recent advances in generative models have enabled the creation of highly realistic counterfeit content, making it readily accessible to anyone armed with nothing more than a smartphone. The prevalence of open-source pipelines and social-media filters at scale today enables non-technical individuals to produce or fabricate very convincing samples, whether in the form of deepfakes or morphing attacks.  Consequently, their reliable detection became a core research challenge at the intersection of computer vision and multimedia forensics.
Early research utilized hand-crafted cues (e.g., illumination or sensor noise inconsistencies \cite{johnson2005exposing, cozzolino2019noiseprint}), but recent studies focus their attention on deep convolutional neural networks (CNNs) that learn discriminative artifacts directly from data. 

In deepfake detection, the dominant pipeline is represented by the fine-tuning of models pre-trained on ImageNet \cite{5206848} such as Xception, EfficientNet, VGG, etc. Even though they achieve near-perfect accuracy on intra-dataset scenarios, these detectors struggle to maintain high performance when tested in cross-dataset settings \cite{zanardelli2023image}, after the media have undergone compression \cite{ZUBAIR2025129116}, or when filters have been applied \cite{ren2025deepfake}. In particular, in \cite{libourel2024case}, the authors produced a deepfake dataset by running Instagram beautification filters over matched real and deepfake clips from the popular CelebDF dataset, then tested three state-of-the-art passive detectors and human observers. They found that the filters were able to mislead both detectors and people, showing that even everyday beauty filters can let deepfakes slip past current defenses and highlighting the need for filter-aware detection methods. However, the literature lacks in-depth analyses of the effect of such filters used separately on genuine faces and deepfakes. 

Regarding the morphing domain, the recent development of detection systems leverages advances in deep learning as well, aimed at automatically learning artifacts either from raw data or hand-crafted descriptors \cite{scherhag2022face}. However, in this context, more focus must be devoted to the generation pipeline, since it can adopt one or a combination of two distinct approaches: landmark-based and generative-based \cite{9380153}. The former approach involves the application of deformations from the facial landmarks of contributing faces, followed by color fusion. In contrast, the latter utilizes generative approaches based on Generative Adversarial Networks \cite{goodfellow2014generativeadversarialnetworks} (e.g., StyleGAN \cite{karras2019stylebasedgeneratorarchitecturegenerative}, MIPGAN \cite{zhang2021mipgangeneratingstrong}) or diffusion models (e.g., DiffMorpher \cite{zhang2023diffmorpherunleashingcapabilitydiffusion}). Both methods have advantages and criticalities: for instance, it is well known that generative methods tend to have a remarkably realistic visual quality, while sacrificing the biometric imprint of the contributors. In contrast, landmark-based approaches are more effective in preserving the biometric imprint of the contributing individuals. However, in this case, the overall visual quality is compromised by the presence of typical deformation and fusion-related artifacts, such as ghosting \cite{venkatesh2020gangeneratedmorphsthreaten}. Consequently, additional post-processing steps are often necessary to eliminate or reduce these artifacts. In this regard, examples of such approaches can be seen in \cite{seibold2023bettermorphedfaceimages, 6485490, zhou_towards_2022}.

Despite the analysis of post-processing methods aimed at improving the deceiving capability of morphs, as far as we know, no work has been done to analyze the impact of beautification filters on morph attack detection. Therefore, to address the limits in the current literature concerning the potential threat of these filters to digital data integrity verification, this work investigates the influence and potential issues of such post-processing steps in the context of morphs and deepfakes detection. 

\section{Experimental Framework}\label{sec:experimental}
To systematically evaluate the impact of beauty filters on deepfake and morphing attack detection, we applied a smoothing filter with increasing values of the application radius and analyzed its effect on the performance of AlexNet and VGG19, two pre-trained convolutional neural networks widely employed in previous research on deepfake and morphing attack detection \cite{sharma2024systematic, panzino2024evaluating}. Our study was conducted on two benchmark datasets, one containing samples for each of the considered facial manipulations. 

The first dataset is CelebDF \cite{Celeb_DF_cvpr20}, a large-scale benchmark dataset for deepfake detection, containing high-quality forged videos generated using advanced face swap techniques, along with their corresponding real counterparts representing celebrities, for a total of 590 real videos and 5639 deepfake videos. It features diverse 59 subjects, varied lighting conditions, and natural facial expressions, making it a challenging and realistic resource for evaluating deepfake detection models. For instance, this dataset can be employed to simulate application scenarios like social media and public content verification, as well as digital forensics.

The second benchmark dataset, for morphing attack detection, is AMSL \cite{neubert2018extended}, containing both bona fide and synthetically morphed face images based on the Face Research Lab London set (FRLL) \cite{DeBruine2017}, featuring neutral and smiling poses from 102 subjects. It is designed to support the evaluation of biometric systems under realistic morphing scenarios, with controlled image quality and identity blending. For instance, it could be employed for simulating authentication scenarios in security contexts, such as border controls.

Both selected networks were trained on 80\% of the samples in the deepfake or morphing dataset (i.e., 20\% for validation) and evaluated performance on the remaining samples. 
The test images were then progressively smoothed to assess the models' robustness against beautification effects, using a filter with increasing application of radius values $c$, ranging from 3\% to 5\% of the face height (\textit{e.g.}, Figure \ref{fig:filters}).

For each model, we report the Equal Error Rate (EER) on the original test samples and after progressive beautification, as well as the Bona Fide Presentation Classification Error Rate (BPCER also known as false positive rate) and Attack Presentation Classification Error Rate (APCER, also known as false negative rate), using the same threshold obtained for the original samples. Finally, we also analyze the related Area Under the ROC Curve (AUC) and the distributions of the scores obtained by the two detectors on real and fake images to provide further insights.

\section{Results}\label{sec:results}

This section presents and discusses the impact of the beautification filters on the presentation attack detection, analyzing separately the effects of this alteration on deepfakes (Section \ref{subsec:df_results}) and morphing attacks (Section \ref{subsec:ma_results}).

\subsection{Deepfake Detection}\label{subsec:df_results}
In the deepfake detection scenario (Table \ref{tab:deepfake}), both networks exhibit a gradual increase in EER as the beautification intensity increases. For instance, AlexNet's EER rises from 22.3\% (original) to 28.1\% at the highest smoothing level, while VGG19 goes from 30.2\% to 35.2\%. 
However, this degradation is driven by two different trends in the two networks. Specifically, through AlexNet it is possible to observe an expected increase in the BPCER, indicating an increasing inability to correctly classify bona fide samples as the beautification becomes more pronounced. The outcomes provided by the VGG show instead that the decay in performance is mainly driven by the increase in the APCER and, therefore, the probability of unrecognized attacks.

\begin{table}[t]
\renewcommand{\arraystretch}{1.3}
\setlength{\tabcolsep}{4pt}
\centering
\caption{Results related to the deepfake detection scenario. AlexNet and VGG19 were trained on the CelebDF dataset \cite{Celeb_DF_cvpr20} and tested on both original images and their smoothed versions obtained through various smoothing radii ($c$).}
\label{tab:deepfake}
\resizebox{\columnwidth}{!}{ 
\begin{tabular}{|c|c c|c c|c c|}
\hline
\multicolumn{1}{|c|}{\multirow{2}{*}{\textbf{Test set}}} & \multicolumn{2}{c|}{\textbf{EER (\%)}}                             & \multicolumn{2}{c|}{\textbf{BPCER (\%)}}                                    & \multicolumn{2}{c|}{\textbf{APCER (\%)}}                                    \\ \cline{2-7} 
\multicolumn{1}{|c|}{}                                   & \multicolumn{1}{c}{\textbf{AlexNet}} & \multicolumn{1}{c|}{\textbf{VGG19}} & \multicolumn{1}{c}{\textbf{AlexNet}} & \multicolumn{1}{c|}{\textbf{VGG19}} & \multicolumn{1}{c}{\textbf{AlexNet}} & \multicolumn{1}{c|}{\textbf{VGG19}} \\ \hline
\textbf{Original} & 22.3 & 30.2 & 22.4 & 30.2 & 22.3 & 30.1 \\ \hline
\textbf{Beautified c = 3.0\%} & 23.0 & 31.0 & 30.4 & 23.0 & 14.3 & 42.7 \\ \hline
\textbf{Beautified c = 3.5\%} & 24.1 & 32.4 & 38.2 & 22.7 & 12.1 & 46.5 \\ \hline
\textbf{Beautified c = 4.0\%} & 25.9 & 33.2 & 48.7 & 22.8 & 9.1 & 48.1 \\ \hline
\textbf{Beautified c = 4.5\%} & 26.5 & 34.1 & 51.6 & 23.8 & 8.8 & 48.1 \\ \hline
\textbf{Beautified c = 5.0\%} & 28.1 & 35.2 & 57.0 & 25.0 & 8.0 & 48.4 \\ \hline
\end{tabular}
}
\end{table}

By analyzing the performance when applying the beautification filter to the real images only and to the fake ones only, it is possible to reveal the reasons behind these findings (Table \ref{tab:deepfake_auc}). In particular, applying the beautification filter to the real images causes a decay in performance of the AlexNet compared to the non-filtered ones, while an opposite trend is revealed when the filter is applied to the fake images. In both cases, the impact is proportional to the smoothing radius. As previously observed, the behaviour of the detector based on VGG19 is different, revealing an improvement when the filter is only applied to real images and a degradation when it is applied to deepfakes. However, in both cases, the increase of the smoothing radius tends to reduce the detection performance, following the trend observed by the application of the filter on the samples belonging to both classes.

\begin{table}[t]
\renewcommand{\arraystretch}{1.3}
\setlength{\tabcolsep}{4pt}
\centering
\caption{Area Under the ROC Curve (AUC) [\%]. AlexNet and VGG19 were trained on the CelebDF dataset \cite{Celeb_DF_cvpr20} and tested on both original images and their smoothed versions obtained through various smoothing radii ($c$).  O-Real and F-Real are original and beautified real samples, respectively. O-Fake and F-Fake are original and beautified deepfake samples, respectively.}
\label{tab:deepfake_auc}
\resizebox{\columnwidth}{!}{ 
\begin{tabular}{|c|c c|c c|c c|}
\hline
\multicolumn{1}{|c|}{\multirow{2}{*}{\textbf{Smoothing  Radius}}} & \multicolumn{2}{c|}{\textbf{F-Real vs F-Fake}}     
 & \multicolumn{2}{c|}{\textbf{F-Real vs O-Fake}} & \multicolumn{2}{c|}{\textbf{O-Real vs F-Fake}} \\ \cline{2-7} 
\multicolumn{1}{|c|}{}                                   & \multicolumn{1}{c}{\textbf{AlexNet}} & \multicolumn{1}{c|}{\textbf{VGG19}} & \multicolumn{1}{c}{\textbf{AlexNet}} & \multicolumn{1}{c|}{\textbf{VGG19}} & \multicolumn{1}{c}{\textbf{AlexNet}} & \multicolumn{1}{c|}{\textbf{VGG19}} \\ \hline
\textbf{Original c = 0\%} & 84.1 & 75.7 & \multicolumn{4}{c|}{}  \\ \hline
\textbf{c = 3.0\%} & 82.8 & 75.1 & 79.0 & 80.5 & 86.9 & 69.9 \\ \hline
\textbf{c = 3.5\%} & 80.5 & 73.4 & 74.9 & 80.4 & 87.8 & 68.3 \\ \hline
\textbf{c = 4.0\%} & 76.9 & 72.1 & 68.8 & 80.2 & 88.9 & 67.3  \\ \hline
\textbf{c = 4.5\%} & 76.0 & 71.2 & 67.3 & 79.4 & 89.0 & 67.2 \\ \hline
\textbf{c = 5.0\%} & 73.7 & 70.2 & 63.9 & 78.7 & 89.3 & 67.1 \\ \hline
\end{tabular}
}
\end{table}

\begin{figure}[t]
  \centering
  \subfloat[AlexNet]{%
    \includegraphics[width=\columnwidth]{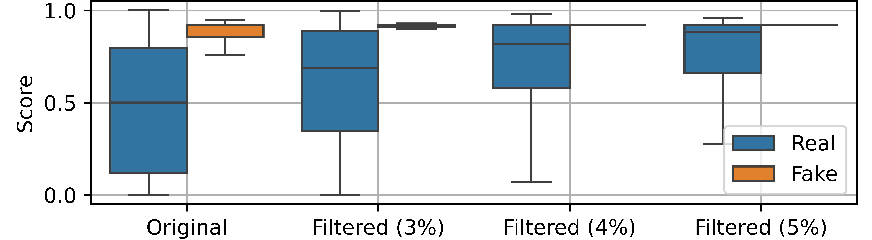}}
  \par\smallskip
  \subfloat[VGG19]{%
    \includegraphics[width=\columnwidth]{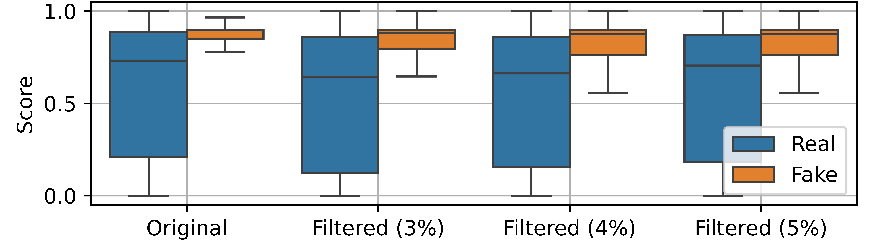}}
  \caption{Scores distribution for real images and deepfakes obtained from AlexNet (a) and VGG19 (b). Scores range from 0 to 1, where the higher the score, the higher the confidence in detecting a deepfake.}
  \label{fig:boxplot_deepfake}
\end{figure}

Figure \ref{fig:boxplot_deepfake} shows that the impact on the stability of the scores is opposite between the two detectors as well. Specifically, while the beautification filter tends to reduce the variability of the scores obtained from AlexNet (Figure \ref{fig:boxplot_deepfake}a) for both real and fake samples, such a filter tends to increase intra-class variability in the case of VGG19 (Figure \ref{fig:boxplot_deepfake}b). Despite the differences in terms of stability, the overall performance decreases after the application of the filter in both cases, due to the shift of the distribution of the scores related to the real samples towards the distribution related to deepfakes in the case of AlexNet, and a greater overlap between the two distributions in the case of VGG19.

These outcomes have important implications and reveal that it is necessary to consider the potential use of beautification filters in deepfake detection, since these may have an unpredictable impact on the performance. In particular, different architectures respond differently to facial manipulations, even if those manipulations are not meant to deceive. For instance, based on the specific deepfake detector, the beautification filters could significantly alter the output, making real images identified as fakes and, more critically, allowing deepfakes to deceive the detection. Therefore, it is necessary to focus on the development of detectors more robust to such subtle, not strictly malicious alterations that could serve as a camouflage for malicious deepfake manipulations.

\subsection{Morph Attack Detection}\label{subsec:ma_results}

In the morphing attack detection scenario (Table \ref{tab:morph}), the performance drops even more significantly. AlexNet’s EER jumps from 27.6\% (original) to 41.2\% at the highest smoothing level, while VGG19’s EER increases from 19.0\% to 37.3\%. 
For both AlexNet and VGG19, such a decay in performance is driven by the BPCER, rising to 70\% and 90\% for c=3\%, respectively, indicating that even minimal smoothing causes genuine faces to be classified as attacks.
This behavior is expected and desirable: in fact, filtering can be considered a manipulation.
On the contrary, APCER decreases drastically, suggesting that fake samples are well-classified more often. 

\begin{table}[t]
\renewcommand{\arraystretch}{1.3} 
\setlength{\tabcolsep}{4pt} 
\centering
\caption{Results related to the morphing attack detection scenario. AlexNet and VGG19 were trained on the AMSL dataset \cite{neubert2018extended} and tested on both original images and their smoothed versions obtained through various smoothing radii ($c$).}
\label{tab:morph}
\resizebox{\columnwidth}{!}{ 
\begin{tabular}{|c|c c|c c|c c|}
\hline
\multicolumn{1}{|c|}{\multirow{2}{*}{\textbf{Test set}}} & \multicolumn{2}{c|}{\textbf{EER (\%)}}                             & \multicolumn{2}{c|}{\textbf{BPCER (\%)}}                                    & \multicolumn{2}{c|}{\textbf{APCER (\%)}}                                    \\ \cline{2-7} 
\multicolumn{1}{|c|}{}                                   & \multicolumn{1}{c}{\textbf{AlexNet}} & \multicolumn{1}{c|}{\textbf{VGG19}} & \multicolumn{1}{c}{\textbf{AlexNet}} & \multicolumn{1}{c|}{\textbf{VGG19}} & \multicolumn{1}{c}{\textbf{AlexNet}} & \multicolumn{1}{c|}{\textbf{VGG19}} \\ \hline
           \textbf{Original} & 27.6 & 19.0 & 25.2 & 18.1 & 30.0 & 20.0 \\ \hline
\textbf{Beautified c = 3.0\%} & 41.2 & 42.9 & 70.0 & 90.0 & 12.4 & 0.0 \\ \hline
\textbf{Beautified c = 3.5\%} & 41.4 & 38.5 & 70.0 & 100.0 & 10.8 & 0.0 \\ \hline
\textbf{Beautified c = 4.0\%} & 40.1 & 40.4 & 70.0 & 100.0 & 9.4 & 0.0 \\ \hline
\textbf{Beautified c = 4.5\%} & 41.1 & 40.1 & 70.0 & 100.0 & 8.0 & 0.0 \\ \hline
\textbf{Beautified c = 5.0\%} & 41.2 & 37.3 & 80.0 & 100.0 & 6.9 & 0.0 \\ \hline
\end{tabular}
} 
\end{table}

The analysis on the impact of individually applying the beautification filter to real images only and to morphed images only confirms the previous results, offering further insights (Table \ref{tab:morph_auc}). Specifically, with both AlexNet and VGG19, the detection performance considerably improves when applying the filter on morphed images only, while the opposite trend is shown when the beautified images are the real ones. This behavior is even more evident when increasing the smoothing radius of the beautification filter.

\begin{table}[t]
\renewcommand{\arraystretch}{1.3}
\setlength{\tabcolsep}{4pt}
\centering
\caption{Area Under the ROC Curve (AUC) [\%]. AlexNet and VGG19 were trained on the AMSL dataset \cite{neubert2018extended} and tested on both original images and their smoothed versions obtained through various smoothing radii ($c$).  O-Real and F-Real are original and beautified real samples, respectively. O-Fake and F-Fake are original and beautified morphed samples, respectively.}
\label{tab:morph_auc}
\resizebox{\columnwidth}{!}{ 
\begin{tabular}{|c|c c|c c|c c|}
\hline
\multicolumn{1}{|c|}{\multirow{2}{*}{\textbf{Smoothing Radius}}} & \multicolumn{2}{c|}{\textbf{F-Real vs F-Fake}}     
 & \multicolumn{2}{c|}{\textbf{F-Real vs O-Fake}} & \multicolumn{2}{c|}{\textbf{O-Real vs F-Fake}} \\ \cline{2-7} 
\multicolumn{1}{|c|}{}                                   & \multicolumn{1}{c}{\textbf{AlexNet}} & \multicolumn{1}{c|}{\textbf{VGG19}} & \multicolumn{1}{c}{\textbf{AlexNet}} & \multicolumn{1}{c|}{\textbf{VGG19}} & \multicolumn{1}{c}{\textbf{AlexNet}} & \multicolumn{1}{c|}{\textbf{VGG19}} \\ \hline
\textbf{c = 0\% (original)} & 75.0 & 87.2 & \multicolumn{4}{c|}{}  \\ \hline
\textbf{c = 3.0\%} & 63.3 & 67.9 & 54.3 & 19.2 & 81.4 & 99.7 \\ \hline
\textbf{c = 3.5\%} & 63.6 & 64.9 & 53.2 & 7.1 & 82.3 & 99.9 \\ \hline
\textbf{c = 4.0\%} & 64.4 & 64.7 & 52.5 & 4.0 & 83.1 & 99.9  \\ \hline
\textbf{c = 4.5\%} & 64.3 & 64.6 & 51.3 & 2.7 & 83.6 & 100.0 \\ \hline
\textbf{c = 5.0\%} & 63.8 & 65.0 & 47.7 & 0.9 & 85.6 & 100.0 \\ \hline
\end{tabular}
}
\end{table}

\begin{figure}[t]
  \centering
  \subfloat[AlexNet]{%
    \includegraphics[width=\columnwidth]{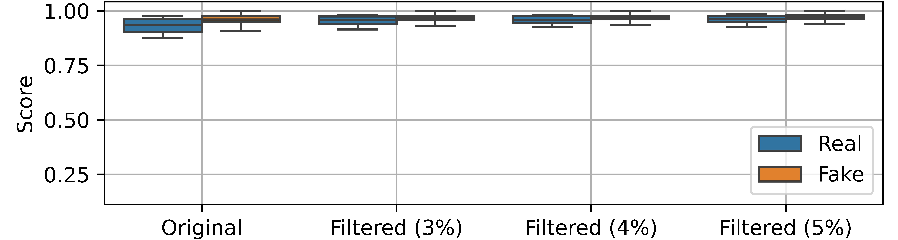}}
  \par\smallskip
  \subfloat[VGG19]{%
    \includegraphics[width=\columnwidth]{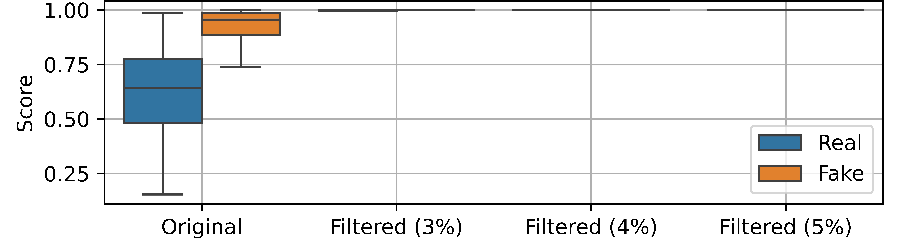}}
  \caption{Scores distribution for real images and morphed images obtained from AlexNet (a) and VGG19 (b). Scores range from 0 to 1, where the higher the score, the higher the confidence in detecting a morphed image.}
  \label{fig:boxplot_morph}
\end{figure}

For both architectures, the beautification filter leads to a reduced variability of the obtained scores, reducing the separation between real and morphed samples compared to the original unfiltered images (Figure \ref{fig:boxplot_morph}). Moreover, the filtering operation causes a shift in the distribution of the scores towards one (i.e., the desirable score for fake samples) for both morphed and real images. This leads to a higher misclassification rate when the filter is only applied to the latter. The beautification filter is more impacting on VGG19 than on AlexNet, making the most performing model on original images the least effective after this unmalicious alteration is introduced. In particular, VGG19 suffers a complete collapse in the detection of bona fide images after even minimal filtering, while AlexNet appears more stable and degrades its performance gradually. An interesting consequence of the trends in the scores is that the application of the beautification filters on only fake images can significantly improve the detection performance, confirming the previous observations (Table \ref{tab:morph_auc}).

To summarize, the beautification filters are able to make the morphed images bypass the detectors and have an even greater impact on performance than deepfake detection on the investigated architectures. Therefore, the potential presence of these alterations in the images could represent a security issue, especially in border controls, and must be addressed in further research.

\section{Discussion and Conclusions}\label{sec:conclusions}

This study highlights the dual impact of facial beautification filters on manipulation detection systems, particularly in the context of digital skin smoothing on deepfake and morphing attack detection. On one hand, these filters are widely adopted by users for non-malicious purposes, such as enhancing aesthetic appearance for social media sharing. On the other hand, the same filters can be deliberately exploited by attackers to conceal manipulation artifacts thanks to the introduced alterations, ultimately undermining the reliability of state-of-the-art detectors.

Our experimental findings on two deep learning architectures, often employed as baseline detectors, demonstrate that both scenarios lead to a degradation in performance. However, the mechanisms vary depending on the network architecture and the type of manipulation. 
Specifically, filters could either reduce inter-class score variability, causing a collapse in score separation, and thus blur decision boundaries or increase intra-class variability, increasing the missclassification ratio due to the greater overlap between the score distributions related to real and manipulated images. These changes weaken the models’ ability to distinguish between real and manipulated inputs, sometimes even adversely impacting more on the effectiveness of originally robust detectors.

\begin{figure}[t]
    \centering
    \subfloat[Original real image, classified as bona fide]{%
        \includegraphics[width=0.45\columnwidth]{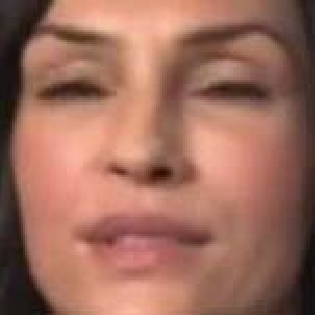}}
    \hfill
    \subfloat[Filtered real image, classified as attack]{%
        \includegraphics[width=0.45\columnwidth]{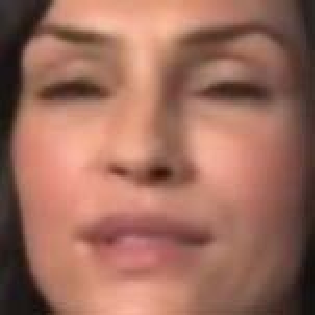}}

    \par\smallskip

      \subfloat[Original deepfake image, classified as attack]{%
    \includegraphics[width=0.45\columnwidth]{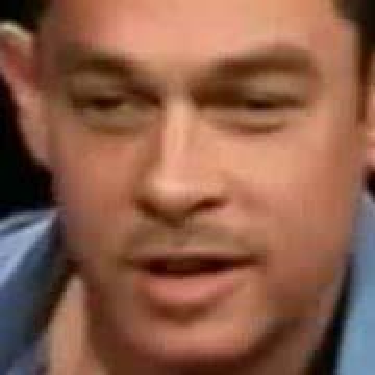}}\hfill
  \subfloat[Filtered deepfake image, classified as bona fide]{%
    \includegraphics[width=0.45\columnwidth]{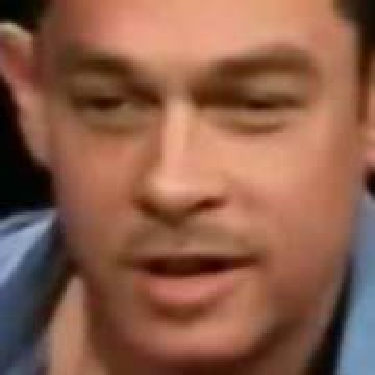}}

  \subfloat[Original morphed image, classified as attack]{%
    \includegraphics[width=0.45\columnwidth]{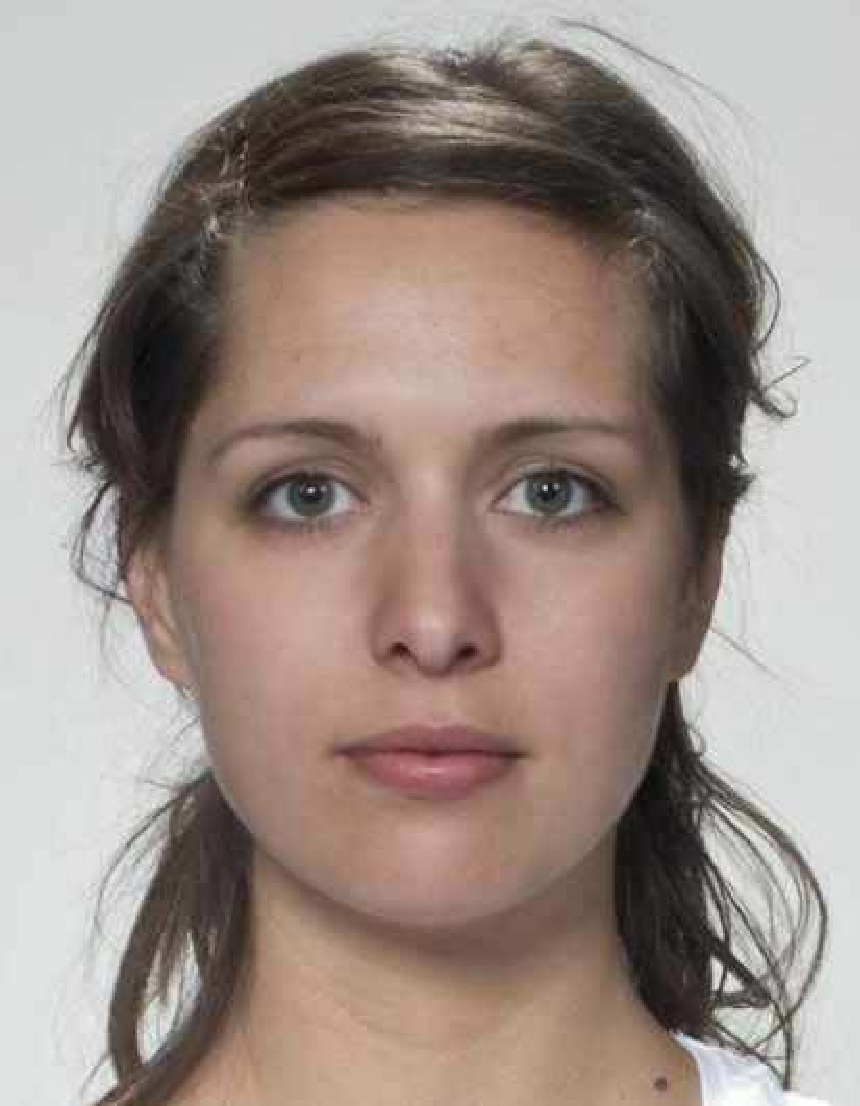}}\hfill
  \subfloat[Filtered morphed image, classified as bona fide]{%
    \includegraphics[width=0.45\columnwidth]{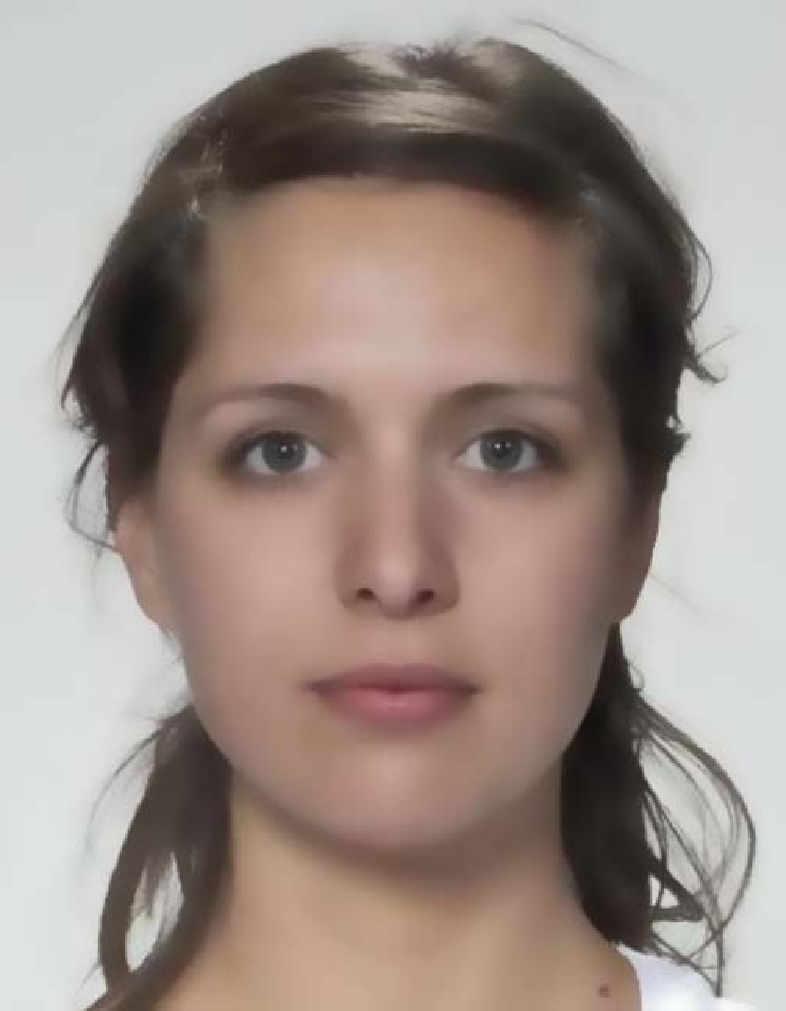}}

  \caption{Effect of beautification on presentation attack detection using the minimum smoothing radius ($3\%$ of the face height). Each row shows an original sample and its filtered counterpart, with the prediction by the AlexNet detector. The first and second rows use samples from~\cite{Celeb_DF_cvpr20}; the third row uses a morph from~\cite{neubert2018extended}.}
  \label{fig:examples}
\end{figure}

Importantly, this degradation is not only present when filters are applied to both real and fake images simultaneously, but is also evident when applied asymmetrically. This suggests that beautification filters can improve the capability of manipulated data in deceiving specific detection systems, either aimed to recognize deepfakes (e.g., Figure \ref{fig:examples}c,d) or morph attacks (e.g., Figure \ref{fig:examples}e,f).
Moreover, even innocent beautification practices by bona fide users can inadvertently resemble adversarial manipulations, leading to increased false rejections (e.g., Figure~\ref{fig:examples}a,b).

Since the impact of such a filtering operation on detection performance strongly depends on the underlying system, various approaches could be investigated by the research community to address this issue. The most straightforward approach is the introduction of beautification filters as an augmentation step to make the system more robust to smoothing operation \cite{bondi2020training}. Considering the different behavior of the individual systems, another possible solution is the joint use of detectors capable of analyzing videos in different ways, therefore leveraging the complementarity of multiple classifier systems \cite{concas2022analysis, la2025exploiting, scherhag2018morph}. A straight attempt could be the combination of models that mutually specialize in filtered and unfiltered input. Another approach is the development of detection models that focus on features that are manipulation invariant or that can be easily adapted to the available data, such as physiological signals \cite{hernandez2022deepfakes} and quality assessment metrics \cite{concas2024quality}, respectively.

In summary, beautification filters pose a threat to the integrity of biometric authentication and forensic analysis systems, making robust deepfake and morph attack detection under such conditions a critical open challenge. Future work should prioritize the development of digital manipulation detection systems that are robust to such subtle, real-world alterations, whether malicious or not, to ensure reliable identity recognition and content verification in everyday and security-critical contexts.

\section*{Acknowledgment}

This work was supported by Project SERICS (PE00000014) under the NRRP MUR program funded by the EU - NGEU and by the PRIN 2022 PNRR  -  BullyBuster 2 – the ongoing fight against bullying and cyberbullying with the help of artificial intelligence for the human wellbeing (CUP: P2022K39K8).


\end{document}